# Lane detection in complex scenes based on end-to-end neural network

1st Wenbo Liu
*School of Information Science and Technology*
*Southwest Jiaotong University*
Chengdu, China
lwb6285658@yeah.net

2nd Fei Yan
*School of Information Science and Technology*
*Southwest Jiaotong University*
Chengdu, China
fyan@swjtu.edu.cn

3rd Kuan Tang
*School of Information Science and Technology*
*Southwest Jiaotong University*
Chengdu, China
ktang_1996@163.com

4th Jiyong Zhang
*School of Information Science and Technology*
*Southwest Jiaotong University*
Chengdu, China
zhangjiyong-com@163.com

5th Tao Deng
*School of Information Science and Technology*
*Southwest Jiaotong University*
Chengdu, China
tdeng@swjtu.edu.cn

*Abstract*—The lane detection is a key problem to solve the division of derivable areas in unmanned driving, and the detection accuracy of lane lines plays an important role in the decision-making of vehicle driving. Scenes faced by vehicles in daily driving are relatively complex. Bright light, insufficient light, and crowded vehicles will bring varying degrees of difficulty to lane detection. So we combine the advantages of spatial convolution in spatial information processing and the efficiency of ERFNet in semantic segmentation, propose an end-to-end network to lane detection in a variety of complex scenes. And we design the information exchange block by combining spatial convolution and dilated convolution, which plays a great role in understanding detailed information. Finally, our network was tested on the CULane database and its F1-measure with IOU threshold of 0.5 can reach 71.9%.

## I. INTRODUCTION

The purpose of lane detection is to help the vehicle find a suitable drivable area and help vehicle make decision, such as avoiding obstacles, changing lanes and overtaking other cars, etc. Therefore, we need algorithms that can understand complex driving scenarios in the lane detection process. Lane line recognition in real scenes is a great challenge. For lane detection in the driverless field, our definition of lane line not only needs to be a normal lane line, but also must include fences, steps and other lines that clearly restrict the driving area. In the simple case where the lane line is very clear and the light is sufficient, the lane detection cannot meet the needs of unmanned driving.

Early the lane detection work was mainly to assist driving, so the scenes faced were relatively simple [1-4]. With the rapid development of deep learning, more and more people propose some methods based on deep convolutional networks to solve the problem of lane line detection. Pan X et al. [5] propose Spatial CNN can more efficiently learn the spatial relationship of feature maps and the smooth, continuous prior of lane lines in the traffic scenario. Spatial CNN uses residuals for information transfer, and their experimental results can prove that the message transfer of Spatial CNN is better than that based on LSTM. Hou Y et al. [6] propose a Self Attention Distillation(SAD) network model, which can allow the network to learn from itself without additional supervision or labels. The SAD module is added to the training, not only can modify the attention map of the shallow block and extract richer contextual information, but the better features learned by the shallow block will affect the deep block, making the final result more accurate. Ghafoorian M et al. [7] propose a network framework using the Generative Adversarial Networks（GAN） to identify lane lines. Send the ground truth to the discriminative model, and the original image send to the generative model; the trained generative model will generate a lane line prediction map. Wang Z et al. [8] and others propose a LaneNet network framework; the decoder in LaneNet is divided into two branches, the embedding branch and segmentation branch. Finally, the results of the two branches can be combined to obtain the result of instance segmentation. Neven D et al. [9] designed a HNet to learn the transformation matrix of the Aerial View transformation under LaneNet's network structure, which can cope with changes in the number and shape of lane lines. Ko Y et al. [10] propose a lane detection method for multi-lane detection based on the deep learning method, and the architecture of the method can reduce some error points generated by the network, thereby improving network performance.

To summarize, our main work is:

- Combine vertical spatial convolution and dilated convolution for lane detection. We design a large number of experiments to get the best combination structure of information exchange block. It can enhance the transmission of detail information in the vertical space of the feature map in the encoding network. This makes the information exchange block designed by us can provide more useful feature information for decoding network in the face of many complex scenes.

- Design feature merge blocks for the non-bottleneck parts of the network. It can promote local network information movement and prevent the loss of detailed information. It enables the whole network to have more detailed information, and the information exchange block enables the network to process the information better. By using it twice in the network encoding part, the accuracy of the lane detection algorithm in complex scenes can be further improved.

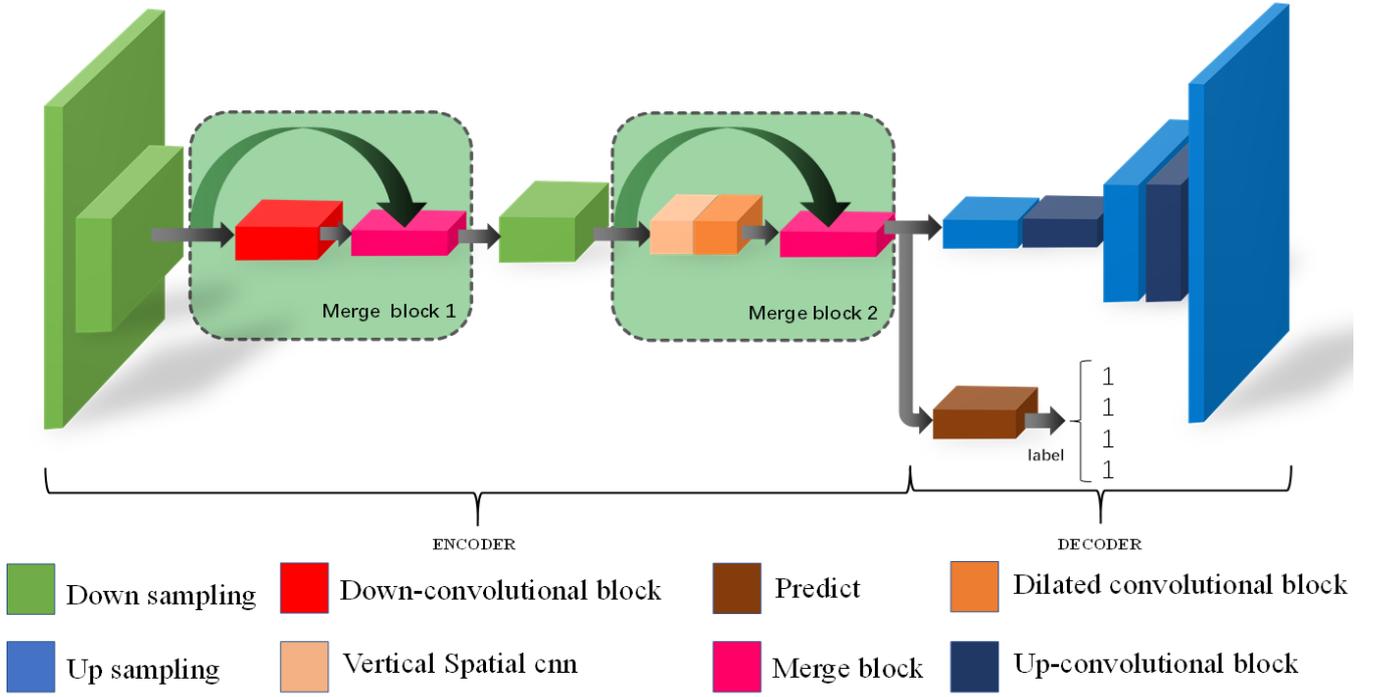

Fig. 1 Lane detection network framework.

## II. NETWORK STRUCTURE

In this section, we discuss the network structure of the network we designed to detect lane lines. Its main frame is shown in Figure 1, which is a single encoder-double decoder structure. We use the end-to-end network structure and the advantage of an end-to-end neural network is that it can directly obtain the required results based on the input data, reducing the data preprocessing, feature extraction and other steps. There is a certain space on the network to fit the problem that needs to be solved. The role of the encoding network part is to reduce the amount of calculation and make the input picture into a feature map understood by the neural network. The function of the two decoding network parts is to predict the existence of the lane line and the probability map of the lane lines. Next, we will introduce the entire network framework from these three parts of the network (A. Encoder B. Predict the probability map C. Predict the existence of the lane line).

### A. Encoder

As shown in Figure 1, it is our overall network structure. The encoding part mainly uses three downsampling to reduce the size of the feature map and increase the channel. The feature map obtained in this way has a larger receptive field [16]. For the detailed of the coding network framework design, we will focus on three parts, namely downsampling, information exchange block and feature merge block.

*1) Downsampling*

Since the two decoding networks share the same encoding structure, the quality of the encoding network plays a significant role in the completion of the lane detection task. Generally, the downsampling operation is completed in the encoding network part, and direct downsampling is easy to lose some details, which is not conducive to our decoding network to directly upsampling.

When performing downsampling like ENet[11], the convolutional downsampling feature map and pooled feature map are stitched together to form a new feature map to save the detailed information well. Our network framework is designed based on ERFNet[12]. It uses the downsampling idea of ENet, as shown in Figure 2, and it is good for preventing the loss of detailed information.

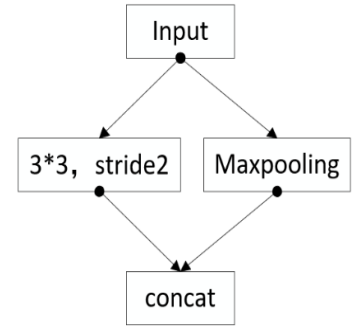

Fig. 2 The downsampling block.

*2) The feature merging block*

The feature merging module used in our network is shown in Figure 3. We design this module mainly for more detailed information to be transmitted to the following network. Only when enough useful information is passed to the network behind, can the network have a better understanding. Another effect of our feature merge block is to prevent the values in the two feature maps merged together from being simply accumulated and sent to the subsequent network. To a certain extent, this block plays a role in preventing errors from being amplified. Through the training of the network, the suitable merging method can be found to retain useful detailed information.

Next we discuss the detailed of the design feature merge block. The role of the feature merging block is to merge the feature map that has passed the middle block with the feature map that has just been down-sampled and reduce the number of channels. First, a 3 * 3 convolution kernel is used to reduce the dimension, and then the 3 * 1 convolution kernel and 1 * 3 convolution kernel are used for feature merging. The idea of using multiple small convolution kernels instead of large convolution kernels is used here, which can reduce parameters and speed up network convergence.

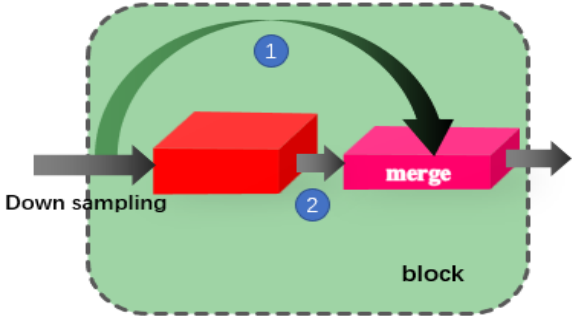

Fig. 3 The feature merging block. It completes the merge of 1 and 2 information flows. It appears twice in the coding network, used with different convolution blocks

*3) The information exchange block*

Our information exchange block consists of vertical spatial convolution and dilated convolution. As we mentioned earlier, the feature merge block allows more information to be passed to the following network, and the information exchange block can better deal with the information. We will introduce the spatial convolution and dilated convolution. The use of dilated convolution is mainly based on the Non-bottleneck-1D block, and the convolution in the Non-bottleneck-1D block is replaced with a suitable dilated convolution. By combining the dilated convolutions of different dilated steps, the effect of dilated convolution is better played[13]. Spatial convolution is equivalent to proposing to train a 3-D kernel tensor K to transmit spatial information in a certain direction, and then pass it in the form of residuals layer by layer. it is verified by experiments that this convolution method can indeed promote the flow of spatial information well. Next, taking the downward spatial convolution(scnn_D) as an example, the formula can be expressed as:

$$x'_{i,j,k} = \begin{cases} x_{i,j,k}, & j = 1 \\ f\left(\sum_m \sum_n x'_{m,j-1,k+n-1} * k_{m,i,n}\right) + x_{i,j,k}, & j = 2,3,\ldots,H \end{cases} \quad (1)$$

where $x_{i,j,k}$ is the element in the i-th channel, j-th row, and k-th column in the feature map, $x'_{i,j,k}$ is obtained after $x_{i,j,k}$ is updated, f is a nonlinear activation function as ReLU[5][14].

Inspired by the principle of spatial convolution, we can not be content to focus only on the relationship between pixels in a small range of the feature map, but also focus on the spatial relationship between pixels in the entire feature map; we must strengthen the transmission of spatial information in the feature map. In the network framework we designed, the information exchange block that uses vertical spatial convolution to transfer information in space layer by layer and dilated convolution can use convolution kernels of different receptive fields, so that feature information can be fully used here understood by the network.

The specific structure of the modules used in the encoding network is shown in Table I below. The Non-bottleneck-1d block[12] is the basic block in our network.

TABLE I. THE SPECIFIC STRUCTURE IN THE ENCODING NETWORK.

| Block | Structure | Depth |
|---|---|---|
| Down-convolutional block | Non-bottleneck-1D*5 | 64 |
| Dilated convolutional block | Non-bottleneck-1D(dilated 1)<br>Non-bottleneck-1D(dilated 2)<br>Non-bottleneck-1D(dilated 1)<br>Non-bottleneck-1D(dilated 4) | 128 |

*B. Predict the probability map*

The task of this decoding branch is mainly to generate the lane line prediction map. Different lane lines of the predicted lane line output at the same time need to be output in different cannels to facilitate the access to the lane line. The resulting lane line prediction map is shown in Figure 4 below. If the prediction map output by the network is a single channel, it is difficult for us to pick the lane lines even if we know the relative position of the predicted lane lines; even if we know the number and relative positions of the lane lines, it will be due to some inaccurate lines. It is difficult to accurately obtain the point on the corresponding lane line. In this branch, we use the decoding structure in ERFNet. It performs upsampling by transposed convolution and Non-bottleneck-1D block.

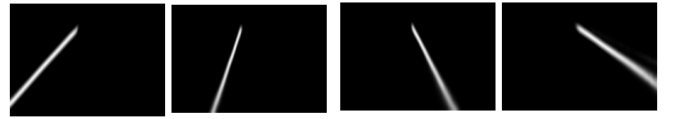

Fig. 4 Predicted lane line probability picture. We can see that different lane lines are predicted into different feature map channels, and some key points are selected and reconnected to obtain the final prediction map.

*C. Predict the existence of the lane line*

As the judgment of the driving area mainly focuses on the lane where the vehicle is located and the adjacent lane, it mainly aims at the prediction of four lane lines in the driving scene. We use the branch predicting the existence of lane lines in the decoding network to predict the probability of the existence of 4 lane lines in the picture.

This branch has two main functions: 1. By weighting the loss of this branch to the loss of the predicted lane line probability map branch, it assists in predicting the lane line probability map. .2 According to the probability of this branch prediction, we can more accurately select some key points of the lane line on the probability map of the lane line prediction to generate a prediction map; the "1 1 1 1" of this branch prediction corresponds to the lane line prediction map predicted by the previous branch, which is very helpful in evaluating the quality of the lane line prediction.

III. EXPERIMENTS

The structure and design details of our network were described earlier, and then we use some comparative experiments to illustrate the effectiveness of our network.

*A. Dataset*

Due to the need for driverlessness , the network used needs to be able to adapt to more complex scenarios, so the experimental selection of this paper was performed under the CULane database. This data set collected a total of 133,235 pictures, divided into 88880 pictures as the training set, 9675 as the verification set, and 34680 as the test set. The corresponding lane detection data in the dataset provides normal scenes and 8 challenging categories. The proportion of each scene is shown in Table II below.

TABLE II. THE PROPORTION OF EACH SCENE

| Scenario | Normal | Crowded | Night | No line | Shadow |
|---|---|---|---|---|---|
| proportion | **27.7%** | 23.4% | 20.3% | 11.7% | 2.7% |
| Scenario | Arrow | Dazzle light | Curve | crossroad | |
| proportion | 2.6% | 1.4% | 1.2% | 9.0% | |

### B. Loss function

First introduce the loss function of the decoding branch for predicting the existence of lane lines. We use Binary Cross Entropy loss, which can well solve the problem of binary classification.

$$Loss_{exist} = -\frac{1}{N}\sum_{i=1}^{N} y_i \log(x_i) + (1-y_i)\log(1-x_i) \quad (2)$$

where $x_i$ denotes the probability that the i-th element is predicted to be a positive sample, $y_i$ denotes the label of the i-th element, N denotes the total number.

The next step for semantic segmentation of lane lines is Cross Entropy loss, then the total loss used by the lane detection decoding branch can be expressed as:

$$Loss = Loss_{ce}(s,s') + 0.1 * Loss_{exist} \quad (3)$$

Where $s'$ denotes the predicted probability map generated by the network, $s$ denotes the ground truth corresponding to the predicted probability map.

### C. Evaluation

We consider the predicted lane line pixel width to 30 pixels as the successfully predicted lane line, and calculate the IOU between the predicted lane line map and its corresponding ground true. Lane lines with a calculated IOU greater than a certain threshold are considered to be True Positive (TP). After getting TP, it is easy to False Positives (FP) and False Negatives (FN) according to the pixel information of the prediction map and the ground truth. According to the formula, the following two indicators can be obtained：

$$Recall = \frac{TP}{TP+FN} \quad (4)$$

$$Precision = \frac{TP}{TP+FP} \quad (5)$$

These two indicators need our comprehensive consideration, so the most common method is to use F1-measure to weigh the relationship between the two indicators. In the end we use as the final evaluation indicator. The final evaluation indicator we used is:

$$F1-measure = \frac{2 Precision\ Recall}{Precision+Recall} \quad (6)$$

Finally, we choose F1-measure with IOU threshold of 0.5 to evaluate the prediction results. In the scene of crossroad, we only compare FP values.

### D. Ablation Study

In this part, we will show the rationality of our network design by comparing some experimental results. We all test the network model with 100 epochs to verify the results. At the same time, the selected IOU refers to the threshold for filtering out the predicted lane line; the higher the IOU threshold, the higher the quality of the predicted lane line.

*1) Compared with similar networks*

Our network model structure is inspired by ERFNet, and it combines vertical space convolution, so we choose three related models to compare with our network. This experiment can prove that our network is not a simple combination of them

TABLE III. RESULTS OF COMPARISON WITH RELATED NETWORK MODELS

| Model | ERFNet+SCNN | Basic SCNN | ERFNet | Proposed |
|---|---|---|---|---|
| F1(0.3) | 79.8 | 80.9 | 80.4 | 80.6 |
| F1(0.5) | 71.0 | 71.6 | 71.8 | 71.9 |

As show in TaBLE III，when the selected IOU threshold is 0.3, the F1-measure of our network is not as high as the data in the original paper of SCNN, but when threshold is 0.5, our model data is higher. The comparison of the data in these experiments proves that our model can effectively integrate the advantages of SCNN into ERFNet, and the information exchange block can further increase the recognition accuracy of the network model in complex scenarios. It also illustrates the points of lane lines predicted by SCNN are more near ground truth than ours, but the quality of our network predictions is higher.

*2) Method used in spatial convolution*

In the process of model design, we tried to use spatial convolution to improve accuracy. The following experiments on how to better use spatial convolution to fit our model.

TABLE IV. THE USE OF DIFFERENT FORMS OF SPATIAL CONVOLUTION

| Model | Spatial convolution | Vertical spatial convolution |
|---|---|---|
| F1(0.3) | 78.9 | 80.6 |
| F1(0.5) | 70.5 | 71.9 |

After experimental verification, vertical spatial convolution can better handle the spatial information of the network framework we designed, preventing the overfull information from being misleading to the network. And vertical spatial convolution is more suitable for our network than other forms of spatial convolution. After experiments, we found the best combination of spatial convolution and dilation convolution in the information exchange block.

*3) Comparison of different model*

In the following experiments, we will compare the impact of different models on the CULane database, and use F1-measure to compare lane detection in each case.

TABLE V. COMPARE THE PERFORMANCE OF DIFFERENT ALGORITHMS.

| Category | SCNN | R-101-SAD | R-34-SAD | ENet-SAD | Proposed |
|---|---|---|---|---|---|
| Normal | 90.6 | 90.7 | 89.9 | 90.1 | **91.1** |
| Crowded | 69.7 | **70.0** | 68.5 | 68.8 | 69.8 |
| Night | 66.1 | **66.3** | 64.6 | 66.0 | 66.2 |
| No line | 43.4 | 43.5 | 42.2 | 41.6 | **44.6** |
| Shadow | 66.9 | 67.0 | 67.7 | 65.9 | **68.1** |
| Arrow | 84.1 | 84.4 | 83.8 | 84.0 | **86.4** |
| Dazzle light | 58.5 | 59.9 | 59.9 | 60.2 | **61.5** |
| Curve | 64.4 | 65.7 | **66.0** | 65.7 | 63.9 |
| Crossroad | 1990 | 2052 | **1960** | 1998 | 2678 |
| Total | 71.6 | 71.8 | 70.7 | 70.8 | **71.9** |

From the results in Table V，the normal scenes account for the highest proportion in the CULane dataset[15]. By comparing the above experimental data, we can find that our algorithm is superior to other algorithms in most scenarios. Therefore, our algorithm can not only solve the problem of lane detection in normal scenarios, but also has sufficient capabilities in other complex scenarios. At the same time, the number of epochs we train is relatively less than other algorithms, and it is easy to converge; it shows the superiority of the information exchange module and feature merging module we designed to process information.

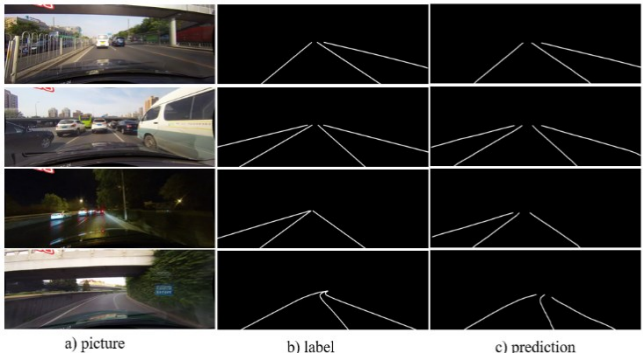

Fig. 5 The results of our proposed network work.

The final result of the lane detection of our network is shown in Figure 5. It can be seen that our method has a good detection effect in a variety of complex traffic scenarios. The first line of the figure is the lane detection under normal conditions. The second line is the detection effect when the lane line is blocked. It can be seen that when the rightmost lane line is substantially blocked by the vehicle, it can still be well recognized by the network. The third line is a rainy night scene. Under such poor light, the middle lane line can still be recognized. The fourth line is the curve. It can be seen that our algorithm still has a good effect in identifying curve lines. We selected the results of four common driving environments for display. We can see that the network we designed can predict the lane line position in a variety of scenarios, which shows that our network structure is reasonable and usable.

## Conclusion

In short, we use vertical space convolution and dilated convolution to design an information exchange module, and use skip-layer connections to implement feature merging modules. After verification, it is found that the new network model can run well in various complex situations, and the modules we added can further improve the accuracy of network identification. Finally, we think that the end-to-end lane line detection method based on deep learning will become more and more attractive, and will be of great significance for the advancement of unmanned driving.